\documentclass{edm_article}
\usepackage[T1]{fontenc} %
\usepackage{fix-cm}      %
\usepackage{amsmath} 
\usepackage{float}
\usepackage{booktabs}
\usepackage{xcolor}
\usepackage{textcomp}    %
\usepackage[pdfpagelabels=false, hypertexnames=false]{hyperref} 
\usepackage{xurl}
\usepackage{microtype}

\newcommand{\revision}[1]{\textcolor{black}{#1}}

\newcommand{\revisioncrc}[1]{\textcolor{black}{#1}}

\DeclareFontSubstitution{TS1}{cmr}{m}{n}

\begin{document}
\title{Temperature and Persona Shape LLM Agent Consensus With Minimal Accuracy Gains in Qualitative Coding}
\numberofauthors{6}
\author{
Conrad Borchers\\
\affaddr{Carnegie Mellon University}\\
\email{cborcher@cs.cmu.edu}
\alignauthor
Bahar Shahrokhian\\
\affaddr{Arizona State University}\\
\email{bshahrok@asu.edu}
\alignauthor
Francesco Balzan\\
\affaddr{University of Bologna}\\
\email{francesco.balzan3@unibo.it}
\and 
\alignauthor
Elham Tajik\\
\affaddr{University at Albany}\\
\email{etajik@albany.edu}
\alignauthor
Sreecharan Sankaranarayanan\\
\affaddr{Extuitive Inc.}\\
\email{sreecharan.primary@gmail.com}
\alignauthor
Sebastian Simon\\
\affaddr{University of Copenhagen}\\
\email{sas@psy.ku.dk}
}
\maketitle
\begin{abstract}
Large Language Models (LLMs) enable new possibilities for qualitative research at scale, including annotation and qualitative coding of educational data. While LLM-based multi-agent systems (MAS) can emulate human coding workflows, their benefits over single LLM agents for coding remain poorly understood. To that end, we conducted an experimental study of how persona and temperature of component agents of a MAS shapes consensus-building and coding accuracy for dialog segments. \revisioncrc{LLMs were prompted to code these segments deductively using a mature codebook with 8 codes and high inter-rater reliability derived from prior research.} Our open-source MAS mirrors deductive human coding through structured agent discussion and consensus arbitration. Using six open-source LLMs (with 3 to 32 billion parameters) and 18 experimental configurations, we analyze over 77,000 coding decisions against a gold-standard dataset of human-annotated transcripts from online math tutoring sessions facilitated by educational software. Temperature significantly impacted whether and when consensus was reached across all six LLMs. MAS with multiple personas (including neutral, assertive, or empathetic) significantly delayed consensus in four out of six LLMs compared to uniform personas. In three of those LLMs, higher temperatures significantly diminished the effects of multiple personas on consensus. However, neither temperature nor persona pairing led to robust improvements in coding accuracy. Single agents matched or outperformed MAS consensus in most conditions. Qualitative analysis of MAS collaboration and coding disagreement may, however, improve codebook design and human-AI coding. 
\end{abstract}

\keywords{large language models, qualitative coding, multi-agent systems, personas, temperature}

\section{Introduction}

Educational research has seen a recent surge of multimodal data produced by increasingly rich forms of user-system interaction \cite{asano2021thematic,pardos2022characterizing,borchers2025augmenting,barany2024chatgpt}. One illustrative example is remote tutoring, where humans and students converse in the process of using educational technology \cite{thomas2024improving,weitekamp2025tutorgym}. These contexts produce transcripts and logs of chat interactions that promise rich insights into the learning process. The surge in rich data stands in contrast with the process of analyzing it: qualitative analysis methods for such data, for example, thematic analysis (TA), requires human skills and domain knowledge and scale poorly with the size of the analyzed dataset. Novel methods of drawing insights from this data are needed, and educational data mining can contribute to the analysis and evaluation of these methods.

To that end, the potential of LLMs for qualitative research has sparked interest among the educational research community. Prior work has demonstrated the promise and limitations of using LLMs for inductive and deductive coding, raising questions about reliability, interpretability, and the degree of human oversight needed. Recent developments in LLM-based multi-agent systems (MAS) offer new possibilities for enhancing the robustness of LLM-based coding, mimicking the collaborative nature of human analysis by distributing roles~\cite{sankaranarayanan2025automating,rasheed2024can,qiao2025thematic}. One key benefit of MAS is that they can implement LLM self-consistency through repeated stochastic generation and aggregation across multiple agents (e.g., re-prompting and majority voting), improving robustness to sampling variance \cite{wang2025mas}. This study builds on that foundation to comprehensively study how temperature and consensus tuning can improve thematic coding accuracy and consensus in educational research settings, if at all.

The contributions from this research are three-fold. First, we introduce a system-centered contribution by developing and empirically evaluating a multi-agent framework for LLM-based deductive coding. The system supports integration of open-source language models and parameter configurations, enabling comparative analyses of their effects on coding reliability, accuracy, and interpretability. Leveraging this framework, we provide a novel empirical comparison of single-agent and multi-agent coding performance aligned with human-coded data, revealing that consensus-making reliably improves coding accuracy only within our sample for a single LLM and code at low temperatures. Second, we systematically evaluate how different LLMs, agent persona congruency, and temperature settings influence consensus-building outcomes using mixed-effects modeling across six models of varying size (3 to 32 billion parameters). This addresses a critical gap in the literature; while persona traits and temperature are theorized to capture multiple perspectives \cite{zhang2023exploring,takata2024spontaneous} and affect the degree to which agents converge on a shared interpretation or consensus \cite{sankaranarayanan2025automating}, their specific impacts on coding dimensions have lacked rigorous evaluation in prior work \cite{li2025llm,simon2025comparing,rasheed2024can,qiao2025thematic}. Third, we present a qualitative analysis of agent deliberation traces to examine how consensus emerges through design choices that closely parallel human-led thematic analysis practices \cite{richards_practical_2017}. By identifying how and when agents revise or reinforce codes during deliberation, this analysis offers deeper insights into the interpretability and epistemic limits of LLM-driven qualitative coding.

To build a methodology for investigating this, we draw from  \cite{barany2024chatgpt}, who employed GPT-4 in a hybrid collaboration with human coders to co-construct codebooks for analyzing tutoring transcripts in math learning facilitated by technology. We extend this work to evaluate how varying system configurations, including different LLMs, persona prompts, and temperature values, shape the thematic outcomes and influence multi-agent coding, focusing on consensus-making and accuracy against human ground truth. Our findings advance theoretical understanding and practical implementation of LLM-assisted qualitative methods by identifying how, when, and why MAS may improve qualitative coding accuracy. Our open-source MAS is compatible with any open-source LLM via Ollama \cite{ollama2023} and produces deliberation traces between agents that are scrutable for researchers and practitioners. This work contributes to ongoing efforts in the field to optimize LLM-mediated interpretive tasks through deliberate system design and structured variation in agent behavior. The research questions guiding this study are as follows:

\begin{itemize}
    \item \textbf{RQ1:} How do agent persona and temperature influence the frequency of consensus-building in a multi-agent LLM system for thematic coding? 
    \item \textbf{RQ2:} Does multi-agent coding with different agent personalities and temperature settings improve upon single-agent coding compared to the human baseline?
    \item \textbf{RQ3:} What qualitative patterns of successful consensus-making arise in multi-agent thematic coding?
\end{itemize}

\section{Background}

This section reviews relevant literature on thematic analysis, the automation of qualitative coding with LLMs, and the emergence of MAS to enhance reliability and interpretability in LLM-driven analysis. It outlines the core processes of thematic analysis (foundational to qualitative coding and established research methods), discusses recent studies that apply LLMs to automate coding, and highlights how multi-agent approaches aim to simulate human collaboration to improve consensus and accuracy.

\subsection{Thematic Analysis}

Qualitative methods like TA offer the potential to move beyond quantitative metrics, which often fail to capture the nuanced dimensions that shape learning processes. TA addresses this limitation by uncovering latent themes and patterns through an interpretive and negotiated process among researchers, ultimately leading to the emergence of meaning \cite{braun2006using,braun2021one}. TA can be conducted using either an inductive or deductive approach. Inductive analysis follows a bottom-up approach, in which themes emerge from the data rather than being shaped by pre-existing theories or frameworks. In contrast, deductive analysis takes a top-down approach, guided by predefined theoretical constructs \cite{braun2006using}.

There are different types of thematic analysis, including reflexive thematic analysis, grounded theory, and content analysis. While these approaches are foundational for conducting qualitative research, translating their underlying theoretical frameworks into consistent practice can introduce variability, depending on a range of contextual factors \cite{barany2024chatgpt,liu2016using}. To mitigate these challenges, researchers often adopt strategies such as: (1) ensuring transparency in the analytic process and thoroughly documenting procedures, (2) iteratively refining the codebook, and (3) establishing inter-coder reliability \cite{barany2024chatgpt}. Although these practices enhance the trustworthiness of the findings, they are also time-intensive, particularly when working with large datasets. As a result, there is growing interest in leveraging LLMs to streamline and support the TA process. The following section explores this emerging area in greater detail.

\subsection{Automated Thematic Analysis}

Although semi-automated, qualitative data analysis tools such as \textit{NVivo} \cite{hilal2013using}, \textit{ATLAS.ti} \cite{smit2002atlas}, and \textit{MAXQDA} \cite{kuckartz2019analyzing} support coding efficiency; they rely on manual coding and interpretation, limiting scalability. Recent work has explored the potential of LLMs to automate or semi-automate TA. For instance, \cite{barany2024chatgpt} compared four coding workflows—manual, automated, and two hybrid GPT-4-assisted approaches—on Algebra I tutoring transcripts. Hybrid methods yielded codebooks with greater clarity, mutual exclusivity between categories, and inter-coder agreement than either fully manual or fully automated approaches. Similarly, \cite{liu2025qualitative} evaluated GPT-4 across three educational datasets, testing various prompting strategies (e.g., zero-shot, few-shot, contextual). GPT-4 achieved substantial agreement with human coders ($\kappa > 0.70$ for 25 of 34 constructs), although it struggled with abstract or ill-defined categories. These findings highlight the potential of LLMs for inductive analysis, though a continued need for human oversight remains.

LLMs have also been studied for deductive coding. \cite{zambrano2023ncoder} found GPT-4 outperformed nCoder in recall and provided helpful feedback for construct refinement, though nCoder achieved higher accuracy. Likewise, \cite{ramanathan2025prompt} proposed GROPROE, an iterative prompting method for deductive TA using theory-based codebooks. After 60 prompt refinements, GPT-4 matched human inter-rater agreement (76\%) on student reflections but remained prone to sycophantic bias. As \cite{de2024performing} highlights, interpretive nuance and contextual understanding still require human judgment.

While promising, these approaches often depend on human feedback and small datasets. In contrast, recent work using MAS for TA, such as Thematic-LM \cite{qiao2025thematic}, suggests that distributed LLM agents may support more autonomous and scalable analysis. The following section introduces this and similar MAS-based approaches to thematic analysis.

\subsection{Multi-Agent Thematic Analysis}

Most prior work on LLM-assisted qualitative analysis has relied on single-agent models, meaning LLMs are applied to code data and no other LLM prompt processes or interacts with intermediate LLM output \cite{liu2025qualitative,zambrano2023ncoder,de2024performing}. Past work has identified several limitations of this approach. \cite{qiao2025thematic} highlights three key issues: (1) the need for human oversight to monitor and contextualize LLM outputs; (2) limited perspective diversity, as single agents often mirror the coder's viewpoint; and (3) the absence of iterative refinement, hindering the model's ability to revise earlier codes. 

In response, recent studies have explored multi-agent LLM systems to address these challenges (e.g., \cite{sankaranarayanan2025automating}). For instance, \cite{qiao2025thematic} proposed Thematic-LM, a multi-agent framework assigning distinct roles—such as coder agents, reviewers, and aggregators—to reduce thematic homogenization and enhance representativeness. Extending this approach, \cite{rasheed2024can} demonstrated the effectiveness of MAS across various qualitative methods (e.g., grounded theory, content analysis), reporting an 87\% practitioner satisfaction rate and improvements in both efficiency and scalability.

However, none of these studies have systematically explored how system parameters, such as temperature or agent persona, shape the effectiveness of consensus-building or coding accuracy in MAS. Moreover, while previous work has proposed architectural frameworks for role assignment among agents, there has been little empirical analysis of how these configurations influence the interpretability or consistency of LLM-generated codes. This leaves open critical questions about whether MAS approaches yield more reliable or representative qualitative insights than single-agent models and under what conditions such improvements occur, a gap we address through systematic prompt experimentation.

\subsection{Consensus in Multi-Agent LLM Systems}\label{sec: ConsensusInMAS}

Consensus seeking is a fundamental challenge in MAS, where agents with divergent initial states must converge toward shared solutions. In LLM-based MAS, agents naturally adopt an ``average strategy'' when no explicit consensus instructions are provided, computing the average of all agents' current states \cite{chen2023multi}.
\cite{chen2023multi} have identified several critical factors that affect consensus dynamics in MAS-LLM systems, including the number of agents, agent persona, and temperature settings. For example, they demonstrated that suggestible agents in two-agent configurations can exhibit oscillatory behavior due to mutual influence without resolution. However, increasing the number of agents stabilizes the system through majority-supported states. In two prior studies, mixed-persona systems with heterogeneous agent groups achieved superior performance in complex reasoning and consensus tasks compared to homogeneous groups \cite{chen2023multi,zhang2023exploring}. However, we are not aware of any study that has demonstrated similar effects in qualitative coding. Moreover, these findings reveal that consensus achievement in MAS depends on agent characteristics, system parameters, and task requirements rather than purely on algorithmic design.

\subsection{LLM Persona Research}\label{sec: humanPersonalityinLLMs}
\paragraph{Persona and Personality Traits in LLMs}
With the advent of LLM agents and their ability to mimic human linguistic behavior, many researchers used the same questionnaires designed to understand human traits (stable differences between people), such as the Big Five personality model, with LLM agents \cite{jiang2023personallm}. In this literature, personality is defined as a stable set of dispositions. In contrast, a persona refers to a situationally induced role or style specified through prompting, instructions, or contextual framing (potentially aligned with a certain personality trait). The results of persona research in LLMs have been mixed. Some research has shown that LLM agents can exhibit intrinsic and extrinsic personality traits both lexically and in decision-making or downstream tasks (e.g., \cite{serapio2023personality}, \cite{jiang2023personallm}, and \cite{tommaso2024llms}). For example, \cite{jiang2023personallm} found that GPT-3.5 and GPT-4 responses matched designated Big Five personality profiles when appropriately prompted. On the other hand, growing evidence points to significant instability in LLM persona expressions. For example, \cite{gupta2023self} demonstrated that trivial changes to prompt phrasing or answer ordering can dramatically alter LLM personality tests. These findings reveal that although LLM ``personalities'' color dialogue style and behavioral patterns, the mechanisms underlying these effects remain poorly understood. Chittem et al. \cite{chittem2025sac} further argues that reliably inducing traits in LLMs requires continuous, nuanced control beyond simple instructions. This uncertainty raises critical questions about the extent to which induced personalities impact LLM agent performance and decision-making in real-world applications, such as deductive coding, a gap that our work aims to address.

\paragraph{Multi-Agent Systems and Personality}
More directly relevant to multi-agent collaboration, research by \cite{zhang2023exploring} and \cite{chen2023multi} demonstrated that systems composed of agents with distinct personalities, such as ``stubborn'' and ``suggestible'' agents, outperform homogeneous agent groups in complex reasoning and consensus tasks. In particular, \cite{chen2023multi} showed how personality traits interact with network topology to influence convergence dynamics: stubborn agents tended to lead to consensus outcomes, sometimes producing hierarchical structures or clustering effects, especially in partially connected networks. Therefore, embedding diverse, interacting personalities in multi-agent LLM systems could also shape the consistency, agreement, and accuracy of multi-agent TA systems, though broader evidence of such effects, especially for qualitative coding, is lacking in the literature.
In contrast, when most or all agents are stubborn, consensus becomes significantly harder to achieve regardless of the number of agents, since such agents rarely revise their positions and the likelihood of deadlock increases \cite{chen2023multi}. Our study builds on this foundation by contrasting dominant and permissive LLM personas, and their combinations, in a rigorous experimental setup. \cite{zhang2023exploring} suggested that odd-numbered agent groups outperformed even-numbered ones by avoiding tie situations during consensus. Therefore, each of our experiments introduces a third agent whenever consensus between the first two agents is not achieved.

\section{System Design}

\subsection{Emulating Human Coding}

Our system design follows established qualitative research practices for deductive coding with a predefined codebook. In standard approaches, at least two trained coders independently assign codes to data segments, assess inter-coder reliability, and iteratively refine codes through discussion, with a third coder often serving as an adjudicator in cases of disagreement \cite{barany2024chatgpt,zambrano2023ncoder,borchers2024using,chen2020efficacy}. We replicate this process by deploying multiple independent AI agents to initially code identical data segments without access to one another's coding decisions, thereby preserving analytical independence. For each coding turn, agents were provided with the current text, the LLM conversation history for the data point being coded (up to the current turn), the instructional prompt, and the codebook. Whereas human coders typically receive explicit training to establish a shared understanding of the coding procedure, AI agents were guided by a system prompt that provided the necessary codebook, contextual information, and procedural instructions. Disagreements were resolved by introducing a third, neutral agent who adjudicated them.

\textbf{Inter-Coder Reliability (ICR) assessment} Agreement between coders is assessed using statistical measures to evaluate consistency \cite{cheung2023use,chandler2024computational}. For instance, according to the literature, a Cohen's kappa of 0.61 or higher indicates substantial agreement between the two coders \cite{landis1977measurement}. Until such an agreement is reached, individual inconsistencies are revisited. This step is approximated in our system by determining whether the two agents fully align on their coding to assess consensus among AI agents over individual data segments.

\textbf{Disagreement resolution}. When coding discrepancies arise, human coders typically engage in structured discussions to reach consensus. This process is often facilitated by a third party or formal consensus process, especially if the discussion between agents does not resolve the disagreement. The new party can activate the majority rule by being the odd addition, or might be a more experienced researcher who can overrule the other coders and make the final decision  \cite{o2020intercoder}. To replicate this critical step, we implemented a consensus agent with a balanced, reflective personality that, upon detecting repeated disagreement (in the previous two turns) between the two coders, makes final coding decisions by considering the original coding prompt, each agent's initial classification, and each agent's discussion.
\subsection{Injecting Personality}
\label{sec:methods:inject-perso}

\paragraph{Methodological Foundation and Trait Selection} Building on the foundations established in Sections \ref{sec: ConsensusInMAS} and \ref{sec: humanPersonalityinLLMs}, we follow recent research that demonstrates that LLMs can be effectively prompted to exhibit distinct persona profiles \cite{newsham2025personality}.
Prior work has established that certain dimensions of the Five-Factor Model exert greater influence in group settings than others. \cite{barry1997composition} found that while conscientiousness showed no relationship to processes and outcomes at individual or group levels, the proportion of extraverted members correlated with enhanced task focus and group performance. Similarly, \cite{mount1998five} demonstrated that agreeableness predicted superior interpersonal cooperation and reduced conflict in team environments. Based on these empirical findings, our study focuses on agents with either high extraversion or high agreeableness, alongside one agent maintaining neutral levels across all traits. This design allows systematic examination of how assertiveness versus cooperation influences consensus formation in multi-agent qualitative coding tasks \cite{tran2025multi}.
The selection of these specific traits is further justified by research showing that prompt politeness and assertiveness significantly influence LLM compliance, including in the generation of false information \cite{vinay2025emotional}. In consensus-making scenarios during qualitative coding, dominance or compliance behaviors may yield different outcomes depending on the accuracy of the initial codes. To investigate these dynamics, each personality trait was implemented through prompt engineering and evaluated for its impact on consensus formation, alignment with human codes, and fallback frequency. Here are the personality profiles used in this study:

\textbf{Bold and Dominant but Elaborative:}
    This personality profile, used for coder agents, reflects assertiveness and extraversion while maintaining a willingness to engage in dialogue. It corresponds to the “bold” category in personality type and was used in multi-agent scenarios. \textbf{Empathetic and Open-Minded:}
    A permissive and collaborative profile emphasizing agreeableness and openness. Labeled as “empathetic” in personality type, this profile was also used in multi-agent settings to encourage broad discussion and interpretation. \textbf{Balanced and Reflective:}
    A neutral profile was applied to the single-agent (control condition) and consensus-agent.

\subsection{Large Language Models}
Another challenge in prompting LLM agents to behave in a certain way is that some LLMs tend to have an inherent personality as reported by \cite{pan2023llms} and \cite{la2025open}. These inherent traits can lead to variability in how models respond to induced personality prompts; some may resist the induced traits and revert to their default behavior, while others may behave inconsistently or unpredictably. \cite{serapio2023personality} and \cite{pan2023llms} demonstrated that \textbf{instruction fine-tuned models} exhibit more reliable and consistent personality traits compared to their base (non-fine-tuned) counterparts. Given our focus on understanding the effect of personality injection and practical resource constraints, we prioritized LLM prompt instruction over task- or persona-specific fine-tuning.
To investigate how personality and behavioral consistency vary across different architectures, we evaluated a diverse set of open-source language models spanning parameter sizes and development lineages. 
The models included in our analysis are:

\textbf{LLaMA 3.2 3B} \cite{touvron2023llama}, developed by Meta, is a lightweight decoder-only transformer model designed for efficient inference in resource-constrained settings, while still performing well on reasoning and multilingual tasks. This model is not instruction-fine-tuned by default. \textbf{OpenHermes V2 (7B)} \cite{teknium2023openhermes} is a community-developed model based on the Mistral architecture, fine-tuned for instruction-following. It demonstrates strong performance on multi-turn dialogue and reasoning benchmarks such as MT-Bench and ARC, ranking among the top-performing 7B open models in human\-aligned chat tasks \cite{zheng2023judging}. \textbf{WizardLM 2 (7B)} \cite{xu2023wizardlm} is jointly developed by Microsoft and Alibaba, optimized for complex multi-turn instruction tuning and is particularly effective in multi-step reasoning and contextual dialogue. \textbf{Phi 3 (14B)} \cite{abdin2024phi} is released by Microsoft, utilizes a dense transformer architecture, and is instruction fine-tuned. This model is trained on a compact, highly curated dataset of synthetic educational material; it performs competitively on benchmarks involving reasoning, math, and safety. \textbf{Mistral-small 3.2 (24B)} \cite{mistral2024small} is an upgrade of their Small‑3.1 24B model, optimized for instruction-following, reducing repetition, and improving function calling. This high-capacity model uses grouped-query attention for efficient inference, excelling at multilingual and factual question-answering tasks. \textbf{Deepseek R1 (32B)} \cite{guo2025deepseek} is a large decoder-only model developed by DeepSeek AI, trained on both code and natural language using a mixture-of-data approach. It achieves strong performance across a variety of benchmarks, particularly excelling in reasoning-heavy tasks and long-context understanding.
    
Our system and experimental evaluation code are fully open-source (repo: \cite{llm_ta_consensus_supplement}) and compatible with the Ollama open-source framework for LLMs \cite{ollama2023}. Codebooks and data can be requested from \cite{barany2024chatgpt} to reproduce our analysis results using our code. The same applies to the LLM collaboration data, which often contains verbatim copies of the input data and is therefore under the same data-sharing constraints.

\section{Methods}

\subsection{Study Context}

\subsubsection{Data Set}
The data for this study were originally collected and annotated by \cite{barany2024chatgpt} and shared with the research team of the present study. The mentioned study was conducted using transcripts from high-dosage math tutoring sessions across two schools, in which trained tutors provided personalized, small-group instruction to support students' mathematics learning. The transcripts were drawn from four 60-minute virtual tutoring sessions. The participants were 9th-grade students enrolled in Algebra I, attending high-poverty schools in an urban region of the northeastern United States. The sessions took place between 2022 and 2023. For this study, we analyzed 579 tutor-student dialog segments from three session transcripts. Each segment is treated as an individual data point for model evaluation.

\subsubsection{Codebook Development}\label{sec:codebookdev}

Our annotation scheme builds on the hybrid human-AI codebook developed in \cite{barany2024chatgpt}, which was rated more favorably by human reviewers than exclusively LLM-built codebooks and found to support more reliable hybrid coding. The codebook includes eight high-level categories (see also study repository \cite{llm_ta_consensus_supplement}): \textbf{Greetings, Instructions, Guiding Feedback, Aligning to Prior Knowledge, Understanding/Engagement Tutor, Technical or Logistics,  Encouragement} and \textbf{Time Management}.

\subsection{Multi-Agent System}
\label{mas:methods}

We developed a multi-agent system to simulate and evaluate AI-assisted qualitative deductive coding based on a codebook. The system includes four components: (1) a single-agent coder, (2) a dual-agent discussion module, (3) a consensus agent, and (4) a hybrid human-AI dataset for benchmarking. All agents operate on the same codebook (see Section \ref{sec:codebookdev}), implemented as a labeled dictionary of binary flags for each codebook category. (e.g., ``Tutor: What's good, XX? \{'Greeting': 1, 'Instruction': 0, ...\}''). Each utterance includes a human-generated reference code derived from double-coding and resolution by trained annotators. The codebook is available in the digital appendix \cite{llm_ta_consensus_supplement}).

The \textit{SingleAgentCoding} module simulates individual annotation. It instantiates a single language model with a fixed persona (e.g., neutral, empathetic, or bold; see Section \ref{sec:methods:inject-perso}) and instructs it to assign codes to a given utterance based on the codebook definitions. The dialog segment and the full codebook are included in the user prompt. To simulate inter-annotator discussion, the \textit{DualAgentDiscussion} module employs two distinct agents with personalities. Agents take turns critiquing each other's code using textual reasoning and are prompted to reflect on discrepancies across multiple rounds. In a post-processing routine, the discussion is checked for completeness, and annotations are extracted. The \textit{ConsensusAgent} module finalizes coding by reviewing both agent responses. It is designed to explicitly identify disagreement and generate a resolved label. The consensus agent maintains a balanced, reflective, neutral persona (similar to the single-coding agent in the control condition), emphasizing justification and adherence to the codebook. The agent receives the data segment utterance and both prior agent responses for reference. The full MAS setup and prompts are available in the digital appendix \cite{llm_ta_consensus_supplement}.

\subsection{Experimental Conditions}
\label{sec:conditions}
We conducted a 3 × 2 factorial experiment varying \textit{decoding temperature} (0.0, 0.5, 1.0; ranging from low to high) and \textit{persona pairing} (congruent vs. incongruent). Persona pairing was defined based on whether the two simulated agent personalities (\textit{persona\_a} and \textit{persona\_b}) matched (congruent) or differed (incongruent). The six pairings included three congruent pairs (balanced-balanced, bold-bold, empathetic-empathetic) and three incongruent pairs (bold-empathetic, balanced\-bold, balanced\-empathetic). The rationale for this design was to investigate whether multiperspectivity, operationalized as persona diversity, influences coding accuracy and consensus formation in LLMs. Each of the 18 configurations was run on a shared dataset of 579 utterances. Agents were instantiated with systematically varied personality traits drawn from three archetypes: \textit{neutral}, \textit{dominant}, and \textit{permissive}. We chose these three traits as the politeness or assertiveness of prompts has been shown to influence LLM compliance \cite{vinay2025emotional} (see Section \ref{sec:methods:inject-perso} for more detail).

\paragraph{Implementation Details}
The system uses the Ollama API \cite{ollama2023} to interface with local large language models (e.g., WizardLM 2 7B). Personalities and roles are injected via structured system prompts that emphasize justification and brevity to keep context windows manageable. Experiments are orchestrated in Python. A single experiment took about 2 hours on a standard MacBook Pro with an M1 CPU (except for DeepSeek R1, which was run on larger machinery). Running the MAS with the DeepSeek R1 model for a single data point on an NVIDIA A100 GPU took, on average, about 60 seconds. We computed the \% \textbf{Agreement Rate} for each experiment, label, and condition, which measures the proportion of utterances in which the final system-assigned code matched the human-coded ground-truth label. The system stores outputs from all modules, including raw and expanded codes, intermediate discussion content, and final consensus decisions. The output is a structured CSV file used for analysis and comparison with human-coded data.

\subsection{Data Analysis Methods for RQ1 and RQ2}
\label{sec:method:analysis:quant}

To address RQ1, which examines the impact of persona and temperature on consensus-making, we analyzed how agent persona pairings and decoding temperature influence the type of consensus achieved during dual-agent discussions. We categorized consensus outcomes into three types: (1) first consensus, where both agents agreed from the start; (2) resolved consensus, where initial disagreement was successfully resolved during discussion; and (3) no consensus, where agents failed to converge on a shared label and the consensus agent needed to enforce consensus. Our focus here lies in examining the likelihood of first consensus and no consensus across persona and temperature conditions.

We estimated generalized linear hierarchical mixed-effects modeling with a binomial link function using the lme4 package in R \cite{bates2015fitting}, predicting each outcome type as a function of agent persona congruency (i.e., matched or mismatched) and temperature (0.0, 0.5, or 1.0; see implementation details in Section \ref{mas:methods}), while controlling for random intercepts by text unit. This adjustment for the text unit is necessary because the same data segments are coded multiple times in our experiments and could systematically vary in difficulty, making it harder to reach consensus (e.g., because some texts are more ambiguous than others). We estimated three generalized linear mixed-effects models with a binomial link function to predict the likelihood of three consensus outcomes: initial, delayed, and no consensus, all following the same formula structure (but with different outcomes): 
\begin{multline}
\text{logit} \left( \Pr(\text{Consensus}_i = 1) \right) =\\
\beta_0 + \beta_1 \cdot \text{Congruency}_i + 
\beta_2 \cdot \text{Temperature}_i + \\
\beta_3 \cdot (\text{Congruency}_i \times \text{Temperature}_i) +
u_{\text{text}[i]}
\end{multline}
In both models, $\text{Congruency}_i$ indicates whether the two agents in condition $i$ had matched personality traits, and $\text{Temperature}_i$ refers to the decoding randomness used by the model during generation. The interaction term captures how the effect of congruency varies across temperatures. The term $u_{\text{text}[i]}$ is a random intercept for each text unit, accounting for potential variation in the difficulty or ambiguity of individual utterances. We highlight that we separately repeated this procedure for each language model.

To evaluate whether MAS coding leads to greater alignment with human annotations than single-agent outputs (RQ2), we analyzed 77,273 individual coding decisions (treating assignment of each codebook category as a distinct binary decision). Each decision represented a binary classification across a fixed codebook, annotated by a human coder, a single-agent model, and a final consensus model. For each decision, we computed whether the model's label matched the human label, both for the baseline single-agent in the control condition and for the MAS's final output. We, again, organized experimental conditions along two dimensions: decoding \textit{temperature} (0.0, 0.5, or 1.0) and \textit{agent persona pairing congruency} (incongruent vs. congruent; see Section \ref{sec:methods:inject-perso} for full details on persona pairing). We fit six models separately by LLM and fit a generalized linear mixed-effects model predicting whether the final consensus matched the human label, using temperature, persona group, and their interaction as fixed effects, and including random intercepts for each utterance and code category (since binary coding decisions are treated as separate observations in this model, as opposed to RQ1):
\begin{multline}
\text{logit} \left( \Pr(\text{FinalAgreement}_i = 1) \right) =\\
\beta_0 + \beta_1 \cdot \text{Temperature}_i + 
\beta_2 \cdot \text{Congruency}_i + \\
\beta_3 \cdot (\text{Temperature}_i \times \text{Congruency}_i) +\\
u_{\text{utterance}[i]} + u_{\text{code}[i]}
\end{multline}
As we did not find any significant and robust trends in how our experimentally manipulated dimensions impacted coding accuracy, we conducted an exploratory analysis. For this analysis, we recoded the persona factor into three levels (includes assertive agent, includes empathic agent, or both), ignoring the neutral-neutral pairing. We then tested model experiments individually (6 LLMs x 8 codes x 3 persona pairings x 3 temperatures = 432 contrasts). To evaluate whether final consensus codes are more closely aligned with human annotations than those produced by a single agent, we performed a paired samples $t$-test. This test aims to compare the percentage of correct code between the single agent and MAS across all contrasts. The Benjamini–Hochberg (BH) method was used to control the false discovery rate \cite{benjamini1995controlling}, given the large number of comparisons. We aimed to determine whether any condition, combination of LLMs, code, or experimental parameter would show reliable MAS improvements relative to the single-coder baseline, and whether any patterns emerged across conditions that produced MAS advantages in qualitative coding accuracy.

\subsection{Qualitative Analysis Methods for RQ3}
\label{sec:analysis:qual}

Our system design features the negotiation of codes through conversational turns, similar to how codes would evolve in a human conversation. Beyond this similarity, our system records all conversational turns, which could be considered recordings of code negotiation in the human analogy. Hence, the system offers a level of transparency not typically found in human coder teams, enabling analysis of the rationale behind each piece of code. We took advantage of this level of transparency to analyze \textit{why} MAS configurations could achieve or fail to achieve the quality of human coding.

To understand the mechanisms behind coding shifts that can speak to patterns of successful MAS collaboration in our context (RQ3), we conducted an ad-hoc qualitative analysis of the \textit{Guiding Feedback} category for OpenHermes, which showed statistically reliable improvements from single- to multi-agent coding, aligning more closely with human consensus. Among 9,735 total coded turns where no initial consensus was reached between the two agents, 71 cases involved \textit{Guiding Feedback} being correctly removed during consensus (according to human ground-truth codes), and 79 cases involved it being correctly added. Two independent human coders analyzed these instances in a spreadsheet. The question guiding the coders' analysis was whether any notable patterns of the agents' consensus-making could explain the observed performance improvement. Coders analyzed: (a) how agents justified their coding decisions, (b) the internal consistency between rationale and codes of agents, (c) the external consistency between turns of agents, and (d) whether certain personalities tend to dominate or concede during discussions. Afterwards, they held a meeting to discuss their findings. In cases of disagreement, they reached consensus through discussion and organized such instances under shared themes, such as Hallucination.

\section{Results}

\begin{figure}[ht]
\centering
\includegraphics[width=\linewidth, alt={Bar chart showing consensus rates across models}]{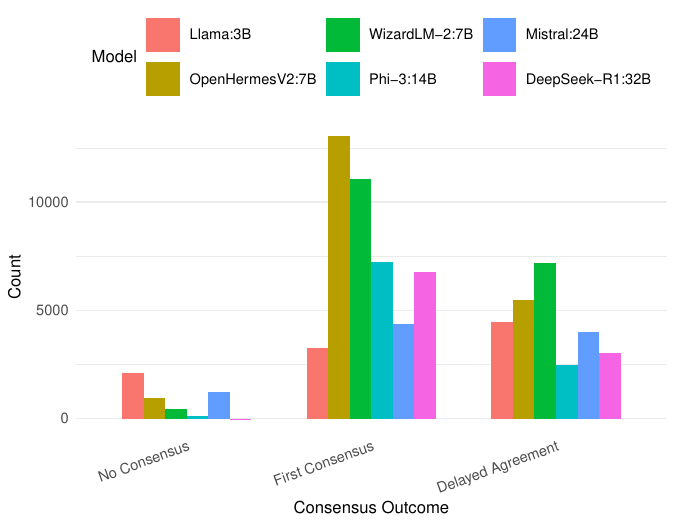}
\caption{Distribution of consensus outcomes over total number of dialog segments and experiments (No Consensus, First Consensus, Delayed Agreement) across the six LLMs.}
\label{fig:descriptive-consensus}
\end{figure}

Figure~\ref{fig:descriptive-consensus} shows the distribution of consensus outcomes across LLMs. Consensus frequencies substantially varied. DeepSeek-R1:32B reached no coding consensus in fewer than 0.1\% of cases, whereas Llama:3b did not reach consensus in 21.7\% of cases. Notably, for all LLMs, immediate consensus in the first round was the most common outcome.

\revisioncrc{The distribution of codes, expressed as percent base rates from human coding, included: Aligning to prior knowledge (6.44\%), Encouragement (5.58\%), Greeting (5.15\%), Guiding feedback (6.44\%), Instruction (9.01\%), Technical or Logistics (0.43\%), Time Management (1.29\%), and Understanding / Engagement--Tutor (26.2\%).}

To investigate how agent persona congruency and temperature influenced consensus-building, we fit a series of generalized linear mixed-effects models with random intercepts for each text. Models predicted three outcomes: (a) first round agreement, (b) no consensus at all, and (c) delayed consensus (agreement only after a second round). Tables in the digital appendix \cite{llm_ta_consensus_supplement} report odds ratios (OR) and 95\% confidence intervals (CI) for each predictor across the three language models. We summarize key findings below.

\subsection{RQ1a: Consensus Making Effects}

\paragraph{Immediate Consensus} 
Across all models, higher temperature significantly reduced the likelihood of reaching consensus in the first round ($p's < .001$), with odds ratios ranging from $\textit{OR} = 0.07$ (OpenHermes-V2:7B) to $\textit{OR} = 0.59$ (DeepSeek-R1:32B). Incongruent personas significantly reduced consensus likelihood in Llama:3B ($\textit{OR} = 0.54$, $p < .001$), WizardLM-2:7B ($\textit{OR} = 0.80$, $p = .010$), Phi-3:14B ($\textit{OR} = 0.67$, $p < .001$), and Mistral-24B ($\textit{OR} = 0.79$, $p = .002$), but not in OpenHermes-V2:7B or DeepSeek-R1:32B. Notably, Llama:3B ($\textit{OR} = 1.96$, $p < .001$) and Phi-3:14B ($\textit{OR} = 1.76$, $p < .001$) showed significant positive interactions between temperature and incongruency. \textit{Full model results are reported in the digital appendix \cite{llm_ta_consensus_supplement}.}

\paragraph{No Consensus} 
Failure to reach consensus was consistently more likely under high temperature across all LLMs ($p's < .01$), with significantly increased odds in Llama:3B ($\textit{OR} = 2.06$, $p < .001$), WizardLM-2:7B ($\textit{OR} = 2.23$, $p = .003$), OpenHermes-V2:7B ($\textit{OR} = 4.15$, $p < .001$), Phi-3:14B ($\textit{OR} = 2.58$, $p = .003$), and Mistral-24B ($\textit{OR} = 3.93$, $p < .001$). In contrast, congruency and its interaction with temperature did not significantly predict no-consensus outcomes in any model ($p$'s $>$ .450). DeepSeek-R1:32B was excluded from this analysis due to an insufficient number of non-consensus cases (less than 0.1\%). \textit{Full model results are reported in Appendix A in the digital appendix (see digital appendix \cite{llm_ta_consensus_supplement}).}

\paragraph{Delayed Agreement} 
Higher temperature significantly increased the odds of delayed consensus across all models ($p's < .001$), with odds ratios ranging from $\textit{OR} = 1.57$ (WizardLM-2:7B) to $\textit{OR} = 10.28$ (OpenHermes-V2:7B). Incongruent personas were significantly associated with increased delayed agreement in Llama:3B ($\textit{OR} = 1.69$, $p < .001$), WizardLM-2:7B ($\textit{OR} = 1.24$, $p = .011$), Phi-3:14B ($\textit{OR} = 1.52$, $p < .001$), and Mistral-24B ($\textit{OR} = 1.21$, $p = .008$), but not in OpenHermes-V2:7B or DeepSeek-R1:32B ($p$’s $>$ .328). Significant negative interactions between congruency and temperature emerged in Llama:3B ($\textit{OR} = 0.52$, $p < .001$), Phi-3:14B ($\textit{OR} = 0.58$, $p < .001$), and Mistral-24B ($\textit{OR} = 0.80$, $p = .040$), suggesting that when generation is more stochastic, the disruptive effects of persona mismatch on delayed agreement are attenuated in some LLMs. \textit{Full model results are reported in the digital appendix \cite{llm_ta_consensus_supplement}).}

Taken together, these findings show that \textbf{temperature robustly delays consensus across all LLMs}, while \textbf{persona congruency exerts more selective effects}. Interaction effects were similarly model-specific, but \textbf{generally showed that higher temperatures attenuate the effects of persona on consensus.} \textit{See digital appendix for detailed model coefficients \cite{llm_ta_consensus_supplement}.}

\subsection{RQ2: Human Alignment}
\label{sec:results:rq2}

Similarly to RQ1, we modeled the likelihood of final alignment with human judgments as a function of temperature, persona diversity (congruent vs. incongruent), and their interaction, using mixed-effects logistic regression with random intercepts for item and code (e.g., greeting vs. praise). Across all six models, persona congruency and its interaction with temperature showed no significant effects on final alignment ($p's $>$ .149)$. Temperature showed a small but significant negative association with alignment in OpenHermes-V2:7B ($\textit{OR} = 0.88$, $p = .003$) and Mistral:24B ($\textit{OR} = 0.85$, $p < .001$), but not in other models ($p's $>$ .158$). Overall, final alignment with human judgments remained largely unaffected by the persona pairing or temperature manipulation, and no single parameter produced reliable improvements in coding accuracy. \textit{See Appendix B for detailed model coefficients in the digital appendix \cite{llm_ta_consensus_supplement}.}

To further explain these findings, Table~\ref{tab:agreement-ci} summarizes the overall percentage of agreement (i.e., alignment with human coders) for each model under both single-agent and multi-agent (final consensus) coding conditions, including 95\% confidence intervals.

\begin{table}[htpb]
\centering
\caption{Percentage agreement (aligned responses) by model with 95\% binomial confidence intervals.}
\label{tab:agreement-ci}
\resizebox{\linewidth}{!}{
\begin{tabular}{lcc}
\toprule
\textbf{Model} & \textbf{Final Consensus (\%)} & \textbf{Single Agent (\%)} \\
\midrule
DeepSeek-R1:32b & 90.0 [89.8, 90.3] & 89.9 [89.6, 90.1] \\
Llama:3B        & 81.6 [81.3, 81.8] & 85.5 [85.2, 85.7] \\
Mistral:24B      & 89.3 [89.1, 89.6] & 89.7 [89.4, 89.9] \\
OpenHermes-V2:7B & 84.9 [84.7, 85.2] & 85.8 [85.5, 86.0] \\
Phi-3 14B       & 89.5 [89.3, 89.8] & 89.6 [89.4, 89.8] \\
WizardLM 2:7B   & 85.6 [85.4, 85.9] & 87.2 [87.0, 87.5] \\
\bottomrule
\end{tabular}
}
\end{table}

\begin{figure*}[htpb]
    \centering
    \includegraphics[width=\linewidth, alt={Difference plot of alignment scores}]{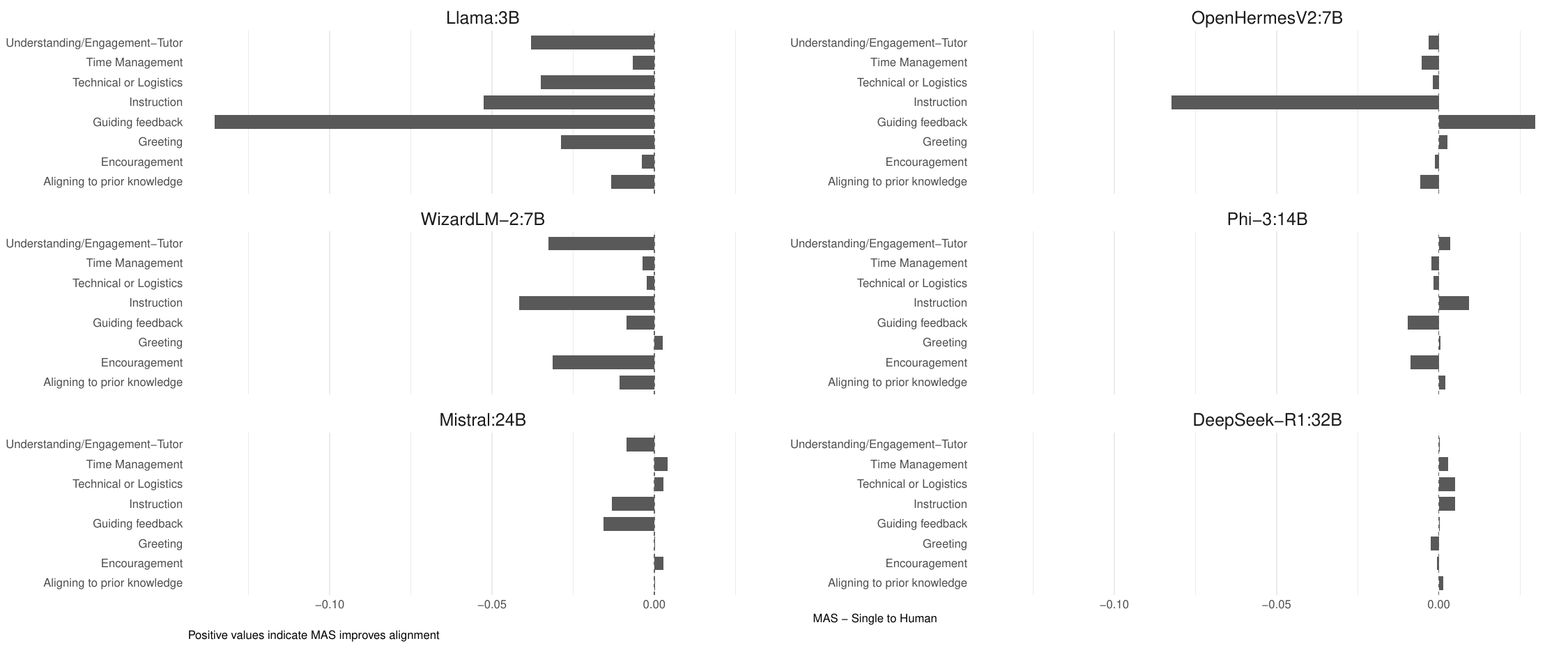}
    \caption{
    Mean difference in alignment between the MAS and single LLM coding with the human labels. Positive values indicate better MAS performance than single LLM coding.
    }
    \label{fig:alignment-diff}
\end{figure*}

Surprisingly, following Table~\ref{tab:agreement-ci}, across all six models, multi-agent coding (i.e., extended deliberation among agents) did not yield substantially better alignment than single-agent coding. In all but one model (DeepSeek-R1:32B), it reduced accuracy, averaged across all experiments. To further examine code-specific effects, we compared alignment between the two coding conditions across different code categories. As shown in Figure \ref{fig:alignment-diff}, some models outperformed others on particular codes, but the pattern of improvement across codes was not consistent across models. Further, the magnitude of performance changes (MAS vs. single agent) also differs across models (notice the difference in y-axis scaling in Figure \ref{fig:alignment-diff} to improve plot readability). For instance, OpenHermes-V2:7B showed improvements of up to 3\% but also performance reductions of about 7\% for the MAS on another code. In contrast, DeepSeek-R1:32B did not show average performance changes greater than 0.5\%. Beyond OpenHermesV2:7B, no model showed an average performance increase of more than 1\%.

The heterogeneity observed in alignment differences raised two questions. First, do comparative differences between multi-agent and single-agent coding vary systematically by item category, temperature, or assigned persona? Second, are the observed differences statistically reliable, or are they false positives due to the large number of comparisons? To address this as part of an exploratory analysis, we applied a modular paired-sample comparison procedure using the Benjamini–Hochberg (BH) method to control the false discovery rate \cite{benjamini1995controlling}. Specifically, we ran paired $t$-tests on agreement scores across coding conditions (multi-agent vs. single-agent), grouped by model, persona, temperature, and code category. We adjusted raw $p$-values using the BH correction across all 432 comparisons (see Section \ref{sec:method:analysis:quant} for more details).

There were 70 statistically significant contrasts after FDR correction. Of these, a minority (5) showed higher agreement in the multi-agent condition, while 65 showed higher agreement in the single-agent condition. All five positive and reliable improvements were on the code ``Guiding Feedback'' with the OpenHermesV2:7B model. In other words, we have no evidence that any other positive gains in Figure \ref{fig:alignment-diff} were due to systematic improvements as opposed to random chance. All of those five included temperatures of 0.5 or lower, and four of those five included bold personalities. Therefore, we conclude that MAS improvements in our dataset were only reliable for a specific model and code. We have no further evidence that consensus-making improves the accuracy of qualitative deductive coding with LLMs.

\subsection{RQ3: Qualitative Insights}

As detailed in the previous section, the OpenHermesV2-7B model showed significantly better alignment with human coders in its multi-agent configuration for the ``Guiding Feedback'' code. Aligned with RQ3 (``What qualitative patterns of successful consensus making arise in multi-agent thematic coding?''), We ran an investigation into the process of 150 coding conversations for the MAS on temperature and persona settings, showing statistically reliable MAS improvements (see Section \ref{sec:results:rq2}). These coding conversations were characterized by either removing the ``Guiding Feedback'' during the conversation, aligning with human coding, or adding the ``Guiding Feedback'' during the conversation, again aligning with the human coding at the end of the MAS conversation. Given that all turns were recorded, we could investigate these successful turns in an exploratory analysis. Two coders each coded 75 conversations in an open coding process, for a total of 900 conversational turns. Notably, we observed that the LLM provided a rationale in only about 80 of 900 conversational turns. In other cases, the LLM explicitly stated that it had no explanation, referenced only a specific data point, or provided no explanation at all. For guiding feedback, the LLM rationale was rare, too, accounting for only 5 of the 80 rationales provided by the agents. Nevertheless, we were able to make several notable qualitative observations about model behavior for this specific subset of the data.

\subsubsection{Consistency}

Beyond the LLM's difficulty in providing a rationale in every turn as instructed, we also observed issues within the reported codes. For instance, agents sometimes incorrectly reported previous coding when introducing it with the phrase ``the previous turn said,'' even though the previous turn's codes did not align with those statements. Other notable conversational turns involved the complete output of the provided codebook without further explication.
Rare hallucinations concerned data points. \revision{In these cases,} two agents discussed non-existent data points.

\subsubsection{New Codes}

In four conversations, the agents proposed new categories. In one example, an agent introduced the new code ``Understanding /  Engagement-Tutor'' to their output, explicitly stating ``The tutor is providing instruction [in this datapoint] on the next step in the math problem while also addressing potential confusion around variables or notation. However, I would like to add subcode under ``Understanding/Engagement-Tutor'' as the tutor is checking for understanding by asking, 'What should we do next?'''.

\subsubsection{Hallucinated Datapoints}
By design, the system analyzes only one data point from the transcript at a time. This means that agents did not have visibility into other codes. Agents sometimes would nevertheless invent and then discuss such data beyond the provided data point. In one instance, in a discussion about the data point ``We found them,'' the second agent mentions in its reasoning that ``the tutor asks if the student remembers what a factor is'', which is not what the data point contains.

\section{Discussion}

This study examined the behavior and performance of a multi-agent LLM system for deductive thematic analysis in an educational setting. Building on prior work on multi-agent qualitative coding, we introduced persona-infused agents. We systematically varied temperature to assess their effects on consensus formation and alignment with human-coded data. The resulting open-source system closely aligns with established qualitative coding practices \cite{zambrano2023ncoder,barany2024chatgpt,borchers2024using}. 

Although existing studies highlight the potential of multi-agent architectures to emulate human collaboration \cite{qiao2025thematic,zhang2023exploring,rasheed2024can}, to our knowledge, no prior work has experimentally evaluated the roles of agent personality, stochasticity, and code-level variation on real-world educational datasets. Beyond empirical evaluation, we considered the implications of these factors for the viability of LLM-based thematic analysis and human–AI collaboration, informed by a qualitative inspection of agent deliberation traces (i.e., justifications for assigning a specific code) to identify conditions under which multi-agent interaction succeeds or breaks down.

\subsection{RQ1: Consensus Building}
RQ1 results showed that decoding temperature had a consistent, statistically significant effect across all models: higher temperatures reduced the likelihood of immediate consensus and increased the likelihood of delayed agreement and no-consensus outcomes. This aligns with prior findings in LLM generation behavior showing that temperature amplifies response diversity and unpredictability \cite{chen2023multi,newsham2025personality}.

We predicted that multiple personalities would lead to \revisioncrc{a more comprehensive consideration of data} during coding. The influence of persona congruency, however, was model-specific. While mismatched (i.e., incongruent) personalities reduced immediate consensus across half of the LLMs we investigated, and delayed consensus in four out of six LLMs. These findings suggest that persona-driven heterogeneity may interact differently with LLM architectures. This finding echoes recent studies arguing that the inclusion of personas to induce multiperspectivity in LLMs is a path full of obstacles \cite{li2025llm}. In particular, findings highlight that the specific relationships between personas and model behavior remain elusive: they may work in some contexts but not others. Therefore, while they might help in surfacing more diverse model behavior in some cases, we have no evidence that they can be systematically and consistently exploited across settings. 

\subsection{RQ2: Human Alignment}

RQ2 assessed whether multi-agent deliberation improves alignment with human-coded ground truth. Overall, multi-agent systems did not consistently outperform single-agent models. Of 432 condition-level comparisons, only five showed statistically significant improvements. In most cases, consensus coding matched or underperformed single-agent performance. These findings align with prior work \cite{li2025single}, which similarly reported higher agreement for a single-agent few-shot approach (82.8\%), compared to a multi-agent setup (79.5\%). \revisioncrc{A practical significance of this finding is that the MAS requires substantially more computational resources (see Section \ref{sec:conditions}), thereby being substantially less resource-efficient for qualitative coding in our context.}

This suggests that extended deliberation among same-model agents does not necessarily yield better accuracy, particularly when the codebook is already well-constructed and disambiguated, as was the case here using a codebook from \cite{barany2024chatgpt}. This finding challenges the assumption that multi-agent deliberation inherently yields greater interpretive robustness and depth, especially when the underlying language model remains constant and deliberative rounds amount to interacting with variants of the same probabilistic system. Despite strong convergent evidence across several experiments and LLMs showing little tangible accuracy improvement, it is possible that better gains for MAS collaboration could be seen in cases where codebooks are less well-developed or when multiple LLMs collaborate. For instance, a resource-efficient approach could be to use smaller models for initial coding, and larger models only when no consensus is reached. Such setups, however, are beyond the scope of the present study. Moreover, it may be that the task and the breadth of coding (i.e., assigning multiple codes per turn) are simply too complex for the LLMs studied here \cite{laban2025llms}.

\subsection{RQ3: Code-Level Variability and Limits}

RQ3 sought to understand where and why MAS may provide additional value in thematic analysis, and what patterns of successful consensus-making emerged. Our observations reveal that, despite the potential of automated systems to assist with qualitative coding, human oversight remains essential for several reasons. First, it could correct technical and formatting inconsistencies, such as invalid code values or schema violations, that undermine data integrity (and can sometimes only be partially mitigated through automated means). In our study, we observed that the only model that reliably improved its own coding in a MAS setup, provided a rationale for coding a data segment in a way that was inconsistent with the instruction in less than 10\% of cases. Second, human coders are needed to identify and rectify logical contradictions, hallucinated justifications, or unsupported code assignments \cite{rawte2023survey}, all of which compromise the quality of automated annotations. This also echoes findings by \cite{laban2025llms}, who demonstrate that LLM performance often degrades in multi-turn conversations, showing increased unreliability and an inability to reproduce earlier instructions compared to fully specified single-turn instructions. While we re-prompted agents up to once when their formatting was incorrect to mitigate this issue, it is possible that performance losses would be smaller if reprompting were performed more often. However, such reprompting also increases computation time and cost.

Further, our analysis revealed a limited  boundary condition under which multi-agent consensus provided its clearest benefit: cases of code ambiguity. Specifically, in the ``Guiding Feedback'' category, consensus agents helped refine overly broad applications of the code and, in some instances, introduced justified reassignments based on context-sensitive deliberation. However, even in these favorable instances, our qualitative inspection revealed significant transparency issues: agent dialogues lacked consistent internal logic, veered erratically, and often failed to provide reasoning that a human could follow or trust. For example, some decisions appeared arbitrary, with agents deferring to previous turns without clear justification or revising positions without explanation. Similar to how recent research has argued that LLM ``reasoning'' is merely an artifact of intermediate token prediction \cite{kambhampati2025stop}, so could agentic collaboration for coding be more akin to improving self-consistency via joint prediction \cite{liang2024internal} rather than a collaborative sensemaking process.

Notably, we observed that coding agents in our experiments sometimes independently proposed new codes, highlighting that consensus mechanisms could benefit human-AI hybrid coding teams \cite{gao2024optimising} in different ways. Rather than maximizing coding accuracy, these systems may be better suited to enhance coding accuracy, codebook development, and challenging existing codes where human coding teams may benefit from new \revisioncrc{interpretations of data}. Our findings thus echo broader calls in the field to retain human expertise in the applications of LLMs for applying expert-level judgment and grading \cite{reza2025prompthive,gao2024optimising}.

\subsection{Implications for Human-AI Coding}

While our system reproduces procedural elements of human workflows (multiple coders, deliberative dialogue, consensus arbitration), it may not reproduce the epistemic stance, contextual grounding, or reflexivity that characterize interpretive qualitative work \cite{braun2021one}. Indeed, our qualitative analysis for RQ3 indicated that agents would often overwrite or change their codes without offering explanations (including, at times, in a contradictory manner), even though they were instructed to provide reasoning to the other agent. This behavior highlights that deductive coding with a codebook is a meaning-making process grounded in the researcher's perspective and iterative engagement with data. Our results suggest that MAS cannot reproduce these epistemic dimensions; nor are they designed to. Despite being framed as reasoning systems, they function as complex retrieval and prediction models that infer the most likely next token as argued by \cite{kambhampati2025stop}.

Findings in this study suggest that the value of LLMs in TA may lie less in reproducing consensus than in disrupting it by introducing ``alternative perspectives.'' Prior work shows that combining human and synthetic data can enrich training outcomes, such as in open-response assessment \cite{borchers2025augmenting}. Our findings similarly support calls for hybrid systems that pair human insight with machine generativity, enhancing qualitative inquiry rather than fully automating interpretive labor \cite{khan2024automating,barany2024chatgpt}. Beyond the manual inspection of agents' consensus processes by humans, data mining methods could also help curate relevant moments in which LLMs question or struggle to disambiguate individual data segments or codes, helping surface analytically ambiguous cases for focused human review \cite{TajikDisagreementAsDataLAK}. Furthermore, detailed analysis of agents' deliberation traces might give better insights into the coding process \cite{li2025single}. In addition, research has shown that LLM outputs can vary significantly depending on how prompts are phrased (e.g., \cite{gupta2023self}), opening yet another perspective for future work.

\subsection{Limitations and Future Work}

Several limitations contextualize these conclusions. First, we relied on a single dataset and a highly refined codebook, previously validated through hybrid human–AI processes \cite{barany2024chatgpt}. The strong performance of the single-agent condition may reflect the maturity and clarity of this codebook \revisioncrc{as well as characteristics of the coding task}. Future work could analyze the impact of codebook size, along with parameters such as the type of training dataset and domain. Second, the system uses multiple instances of the same base model, prompted differently. While personalities were carefully engineered and experimentally compared (as well as theoretically justified; see Section \ref{sec:methods:inject-perso}), the agents were instantiated solely through prompt-level instructions and all rely on the same underlying architecture. Although some research has demonstrated that LLMs tend to prefer the output generated by their own \cite{panickssery2024llm}, it remains an open question whether multi-agent diversity derived from different model families (e.g., mixing Claude, GPT-4, LLaMA), as well as more extended and direct interactions between agents, would lead to even richer or more robust consensus processes. Third, while we focused on deductive coding tasks, future studies might examine inductive coding or mixed-methods settings where codebook refinement and creation are explicitly part of the analytic process \cite{barany2024chatgpt}.\revisioncrc{Finally, given that DeepSeek-R1, the most reasoning-intensive model in our study, did not yield accuracy gains over single-agent coding, future work could examine whether the deliberation traces produced by such models offer utility beyond accuracy, particularly for codebook refinement \cite{TajikDisagreementAsDataLAK}.}

\section{Conclusion}

As educational technologies generate increasingly rich textual data, such as chat and audio transcripts, methods for annotation require careful validation. Large language models have drawn attention for their potential to automate qualitative coding at scale. This study examines how multi-agent configurations of LLMs affect deductive coding processes and outcomes. Although multi-agent systems are often assumed to improve reliability and interpretive depth, our results challenge this assumption. Across experiments with six open-source LLMs, multiple persona pairings, and temperature settings, we find that multi-agent structures strongly shape consensus dynamics, influencing when agents converge or diverge, with persona effects varying across models. However, these changes yielded minimal and highly conditional accuracy gains.

Multi-agent deliberation only marginally outperforms single-agent predictions, and only under narrow conditions involving a specific model, code category, low temperature, and an assertive agent. This calls into question the reliability of agent collaboration as a general-purpose method for improving accuracy or diversity in deductive qualitative coding. Instead, our qualitative analyses suggest that MAS are better suited to exposing ambiguity, prompting codebook revision, and surfacing interpretations that human-only teams might miss. By showing when multi-agent systems do not improve coding, this work helps constrain overgeneralized claims about agentic LLM collaboration.

\bibliographystyle{abbrv}
\bibliography{sigproc}

\balancecolumns
\end{document}